\documentclass[10pt,letterpaper]{article}

\usepackage{cogsci}
\usepackage{amsmath}
\usepackage{amsfonts}
\usepackage[style=apa,sortcites=true,sorting=nyt,backend=biber, uniquename=false]{biblatex}
\cogscifinalcopy 
\usepackage{graphicx}
\usepackage{bbm}
\usepackage{pslatex}
\usepackage{float} 

\usepackage{makecell}

\title{Recovering Mental Representations from Large Language Models \\ with Markov Chain Monte Carlo}
 
\author{{\large \bf Jian-Qiao Zhu (jz5204@princeton.edu)} \\
  Department of Computer Science \\
  Princeton University
  \And {\large \bf Haijiang Yan (haijiang.yan@warwick.ac.uk)} \\
  Department of Psychology \\
  University of Warwick
  \AND {\large \bf Thomas L. Griffiths (tomg@princeton.edu)} \\
  Department of Psychology and Computer Science\\
  Princeton University
  }

\begin{document}

\maketitle

\begin{abstract}
Simulating sampling algorithms with people has proven a useful method for efficiently probing and understanding their mental representations. We propose that the same methods can be used to study the representations of Large Language Models (LLMs). While one can always directly prompt either humans or LLMs to disclose their mental representations introspectively, we show that increased efficiency can be achieved by using LLMs as elements of a sampling algorithm. We explore the extent to which we recover human-like representations when LLMs are interrogated with Direct Sampling and Markov chain Monte Carlo (MCMC). We found a significant increase in efficiency and performance using adaptive sampling algorithms based on MCMC. We also highlight the potential of our method to yield a more general method of conducting Bayesian inference \textit{with} LLMs.

\end{abstract}

\section{Introduction}

How do we know what representations artificial intelligence (AI) systems are using? For ``white box'' machine learning models, such as decision trees and Bayesian models, the representations are typically  transparent and directly tied to the features and arhitecture of the model. Interpreting these models often involves looking at the coefficients, rules, or structures they use to make predictions. However, state-of-the-art AI systems frequently employ ``black box'' deep neural networks (e.g., \cite{lecun2015deep, vaswani2017attention}), which are notoriously difficult to interpret. 

The increasingly proprietary nature of models used in AI can also mean that their internal mechanisms are not readily accessible, posing a significant challenge for researchers who seek to understand the representations used by these models. Historically, the representations used by neural network models have been identified by analyzing the activation patterns of artificial neurons (e.g., \cite{kornblith2019similarity}). However, the efficacy of neuron-level approaches diminishes as AI systems expand in both depth and the number of model parameters. In this context, we propose an alternative approach, drawing inspiration from cognitive psychology, to investigating the  representations used by AI systems via their behaviors (i.e., the outputs they produce).

Cognitive psychologists have spent decades developing methods for elucidating the content of individuals' mental representations, such as the structure of object categories and the utilities assumed to different choice actions \parencite{torgerson1958theory, shepard1979additive, sanborn_uncovering_2010}. These mental representations, while not directly observable, can be inferred through the analysis of behavior. In this paper, we  adapt behavioral methods based on sampling from subjective probability distributions to AI systems. We evaluate the efficiency and performance of three such methods, with the goal of exploring the correspondence between the representations inferred from AI systems and those of humans. 

Our focus in this paper is on recovering color representations of an object, which can be defined within a 3D space. This choice is strategic: it addresses the concern that in simpler domains certain behavioral methods are not distinguishable from each other, while in more complex domains the visualization of results becomes challenging. Formally, an agent's color representation can be conceptualized as a probability distribution over a color space $x$, conditioned upon a given object $c$, expressed as $p(x|c)$. Here, $x$ represents a color defined in terms of Hue, Saturation, and Lightness (HSL) values. For instance, the mental representation of a strawberry’s color would be represented as a probability distribution across a range of colors, each specified by unique HSL parameters.

Our analysis focuses on GPT-4 as an example system, based on its impressive ability to solve a wide range of problems that were previously only solvable by humans. Its capabilities extend to engaging in open-ended dialogues and demonstrating a surprising familiarity with visual concepts \parencite{rathje2023gpt, bubeck2023sparks}. The remainder of this paper is dedicated to applying behavioral methods to extract and analyze GPT-4’s representation of color. It is important to note, however, that the applicability of these behavioral methods is not confined solely to GPT-4. Indeed, these methods can be readily adapted and applied to other AI systems, provided they possess the necessary knowledge base. This flexibility highlights the potential for broader implications and uses of our methodological approach in the evolving landscape of AI research.

\begin{figure*}[t!]
    \centering
    \includegraphics[width=\textwidth]{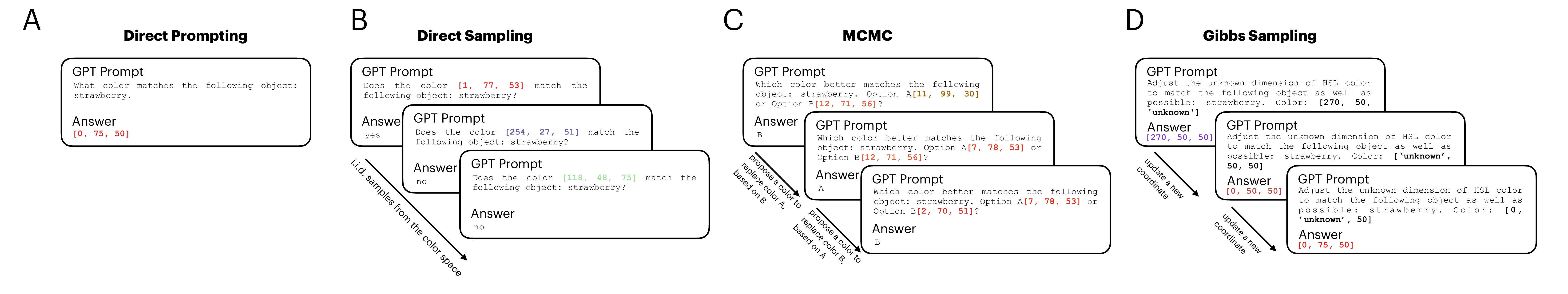}
    \caption{Illustrations of the four behavioral methods used to recover mental representations for GPT-4. 
    \textbf{(A) Direct Prompting with GPT-4:} GPT-4 is directly prompted to generate a HSL color code corresponding to a specified object.
    \textbf{(B) Direct Sampling with GPT-4:} In this iterative process, a random HSL color code is sampled and presented to GPT-4, which then evaluates the extent to which this color matches the target object.
    \textbf{(C) Markov chain Monte Carlo (MCMC) with GPT-4:} Each iteration involves proposing a new color, derived from the previously selected color, and then deciding whether to accept this new color or retain the old one.
    \textbf{(D) Gibbs Sampling with GPT-4:} In each step, GPT-4 is tasked with deducing and filling in a missing dimension of the HSL color code to better match the target object. 
    In all panels, HSL color codes are colorized to assist easier comparison. }
    \label{fig:methods}
\end{figure*}

\section{Background}

Our work draws on a class of cognitive psychology methods that elicit mental representations in humans by integrating people into sampling algorithms. A notable example of such an approach is the World Color Survey \parencite{kay2009world}. In this survey, people are presented with colors that exhaustively sample from the color space, and they are then asked to provide evaluations of these colors \parencite{kay2009world}. However, exhaustively enumerating every possible stimulus quickly becomes infeasible for dealing with high-dimensional or continuous-scale stimuli because the space is simply too vast to explore thoroughly. In contrast, more recent methods that have adopted adaptive sampling algorithms, such as Markov chain Monte Carlo (MCMC), to explore people's representation more efficiently. These methods have shown enhanced efficacy in exploring the structure of mental representations in domains including color, emotional prosody, face, and fruit (e.g., \cite{sanborn_markov_2007}).

\subsection{Probing Large Language Models}
Approaches to probing and interpreting information encoded in LLMs at the neuronal level primarily involve associating internal representations with external properties. This is done by training a secondary classifier on the activations of artificial neurons, with the aim of predicting specific properties \parencite{alain2016understanding}. Researchers typically use a trained LLM to generate representations, then employ another classifier that uses these representations to predict a certain property. This method has shown promise in assessing whether LLMs encode syntactic information \parencite{belinkov2022probing} and, more recently, in analyzing the semantic structure of sentences \parencite{zhang2023deep}. However, we diverge from these methods by focusing on recovering representations from LLMs using behavioral methods, making our work complementary to existing approaches.

\section{From Behaviors to Representations}

As shown in Figure \ref{fig:methods}, we tested four behavioral methods, which can be broadly categorized into two classes: static and adaptive. Static methods typically involve presenting participants with a predefined set of stimuli, selected by the researcher prior to the commencement of the experiment. These methods do not modify the stimuli in response to participants' judgments during the course of the experiment. Examples of static methods include Direct Prompting and Direct Sampling.

In contrast, adaptive methods dynamically tailor the selection of stimuli for participants based on their previous responses. This approach allows for a more dynamic and responsive experimentation process. Notable examples of adaptive methods are Markov chain Monte Carlo (MCMC) and Gibbs Sampling with People \parencite{sanborn_markov_2007, harrison_gibbs_2020}. Both methods iteratively adjust the selection of stimuli, with the aim of achieving a more accurate representation of the participant's mental state by considering their prior judgments.

\subsection{Direct Prompting with GPT-4}
Perhaps the most basic behavioral method to elicit an agent's mental representation involves instructing it to introspectively disclose it. In GPT-4, this could be achieved by directly prompting the model to reveal the conditional probability $p(x|c)$ by providing the object $c$.

For example, in exploring the color representation of a strawberry, we directly prompted  GPT-4 with the following text:
\textit{``You are a participant in a color judgment task. You will be asked to describe an object's color in each question. Your objective is to generate an apt color code in HSL format to match the given object as well as possible. Remember, it’s essential to answer the question with a single HSL code, even if the generated color or the object might seem unusual at times. Please limit your response to just the three values of the HSL code, for example, `h, s, l'. What color matches the following object: strawberry.''}

\begin{figure*}[t!]
    \centering    
    \includegraphics[width=0.95\textwidth]{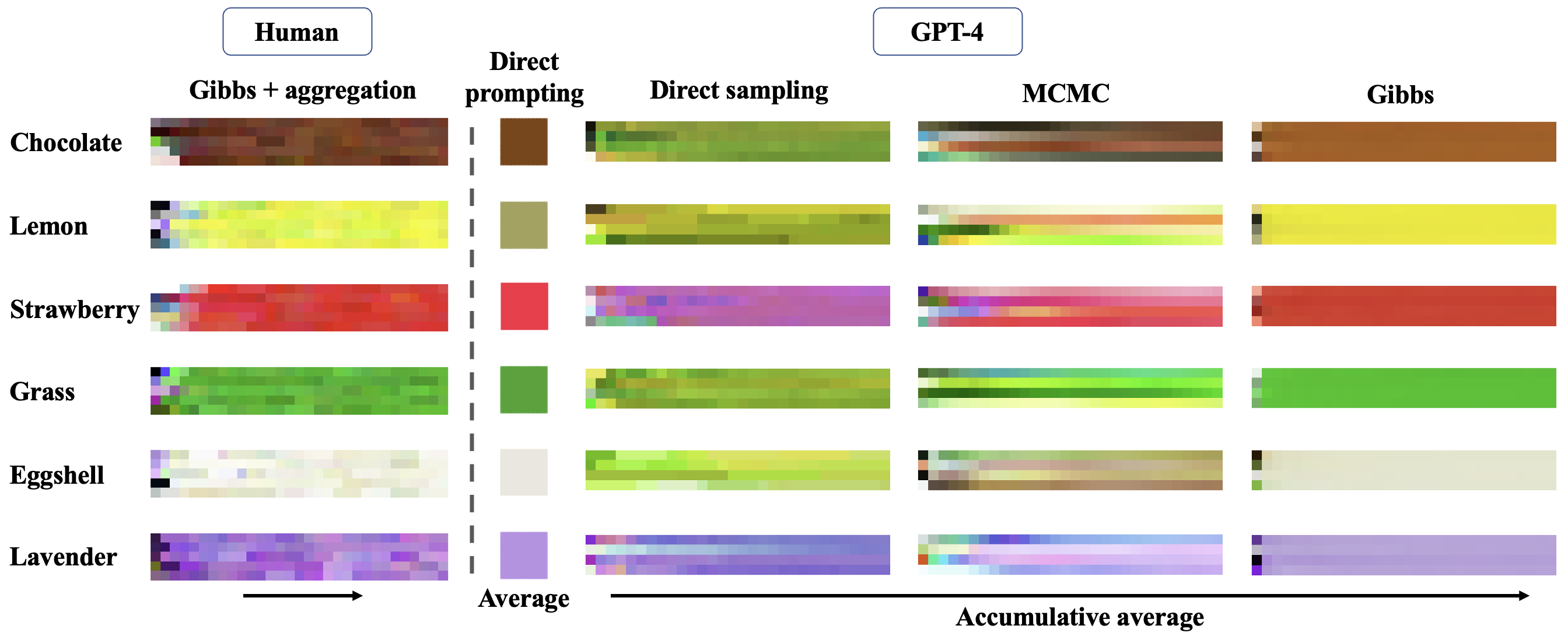}
    \caption{The evolution of the mean color representation across successive iterations. Each row within a color patch represents a single chain. Human data were adapted from \textcite{harrison_gibbs_2020}. }
    \label{fig:sample}
\end{figure*}

\subsection{Direct Sampling with GPT-4}

An alternative static method that circumvents the need for the agent to explicitly report a full color code is Direct Sampling. In this approach, the researcher randomly sample a valid HSL color code at each step, denoted as $x_i \sim p(x)$. $p(x)$ is a uniform distribution over the entire color space. Subsequently, GPT-4 is tasked with determining whether the sampled color corresponds to the specified object, indicated by $\mathbbm{1}_{c}$. That is, GPT-4 is only required to make a binary choice at each step, simplifying the task of directly reporting a color from scratch. Gradually, we approximate the conditional probability with the positive examples that were classified as the object:
\begin{align}
    x_{\mathbbm{1}_{c}} \sim p_c(x) = p(x|c)
\end{align}

Using the same strawberry example, we implemented Direct Sampling with GPT-4 as follows: \textit{``You are a participant in a color judgment task. You will see a question about whether a color (represented in HSL format) matches an object. Simply answer either 'yes' or 'no' based on your interpretation of the object’s color in the question. Does the color [300, 97, 48] match the following object: strawberry?''}

\subsection{Markov Chain Monte Carlo with GPT-4}

MCMC with People (MCMCP) is a well-established adaptive method to elicit people's mental representations \parencite{sanborn_markov_2007, sanborn_uncovering_2010}. We adapted the method to GPT-4. The key idea is to construct a Markov chain whose stationary distribution is $p(x|c)$, and thus the sequence of states generated by this chain can be interpreted as samples from the stationary distribution.

The Markov chain is initiated with an arbitrary value, $x$. To progress the chain, a new candidate value, $x'$, is generated by sampling from a proposal distribution $q(x'|x)$. Then the agent makes a decision on whether to accept $x'$ based on its relative probability compared to $x$ under the target distribution $p(x|c)$. This process hinges on two key assumptions: (i) the proposal distribution is symmetric, $q(x'|x)=q(x|x')$, and (ii) the probability of accepting the proposed value matches the Barker acceptance function \parencite{barkerMonteCarloCalculations1965}:
\begin{align}
    A(x'|x,c) = \frac{p(x'|c)}{p(x'|c)+p(x|c)}
\end{align}
Under these conditions, the sequence of states generated by this Markov chain will converge to a stationary distribution that is consistent with $p(x|c)$.

Here we specify the proposal distribution for 90\% of trials as a multivariate Gaussian distribution with a covariance matrix that is an identity matrix multiplied by 30: $q(x'|x) = N(x, 30 \mathbf{I}_3)$. On the other 10\% of trials, the proposed stimulus was sampled uniformly within the color space, facilitating large jumps in the stimulus space \parencite{martinTestingEfficiencyMarkov2012}. In each iteration, GPT-4 was given a binary choice between two options, $x$ and $x'$. The positions of these options were randomized.

The structure of the prompts implementing MCMC with GPT-4 is as follows: \textit{``You are a participant in a color choice task. You will see a question with two color options in HSL format. Simply choose either Option A or Option B. Remember, it’s essential to pick one color that better matches the object in the question, even if the choices might seem unusual at times. Please limit your response to just 'A' or 'B'. Which color better matches the following object: strawberry. Option A[0, 53, 12] or Option B[274, 81, 47]?''}

\begin{figure*}[t!]
    \centering
    \includegraphics[width=\textwidth]{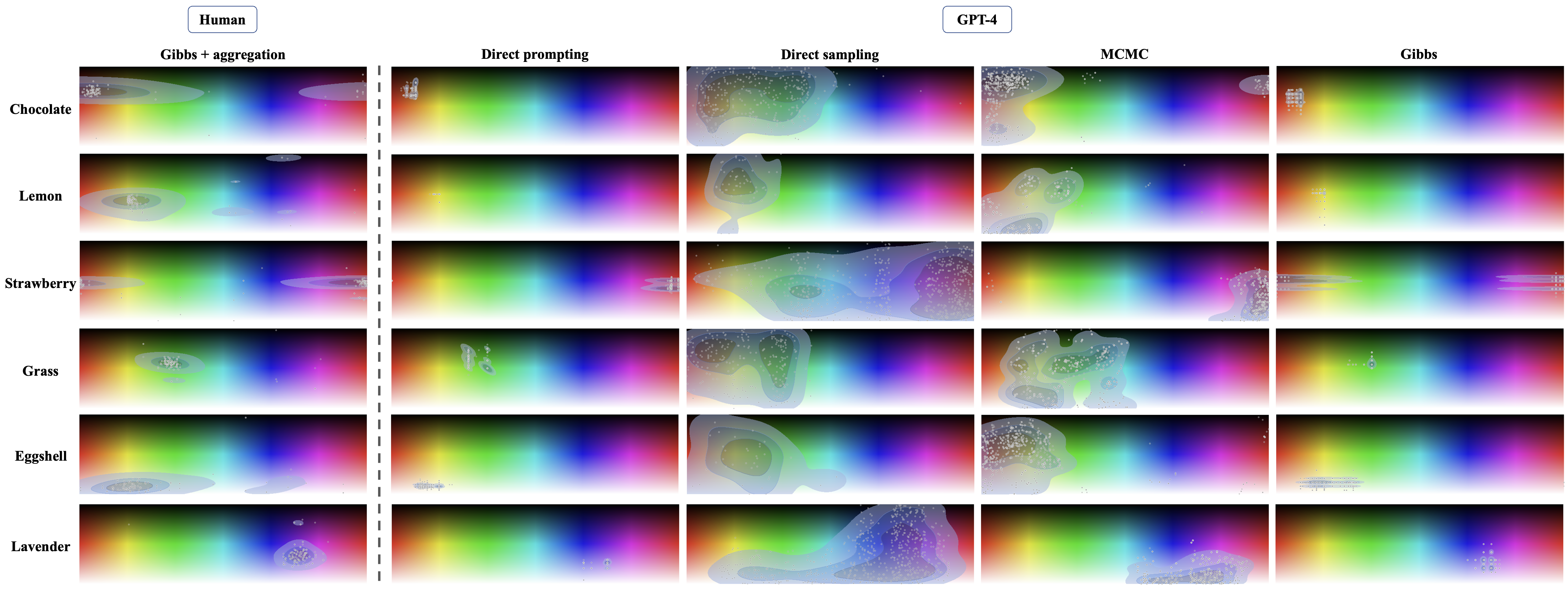}
    \caption{Samples in the color space produced by humans and those generated by GPT-4 using the four behavioral methods (displayed as columns). The overlaid contours are estimates derived from kernel density using a Gaussian kernel with a bandwidth of 1. }
    \label{fig:percept}
\end{figure*}

\subsection{Gibbs Sampling with GPT-4}

Gibbs Sampling with People (GSP) is a recent extension of the MCMCP method \parencite{harrison_gibbs_2020}. Gibbs Sampling involves cyclically sampling from each dimension based on the conditional probability $p(x_{k}|x_{-k},c)$ \parencite{geman1984stochastic}. Analogously, in GSP, participants contribute to the update of coordinates. This is achieved by adjusting a slider corresponding to the current stimulus dimension, $x_k$, while keeping the other dimensions, $x_{-k}$, constant \parencite{harrison_gibbs_2020}. The fundamental assumption in GSP is that participants select an option $i$ for the $k$-th dimension following a specific probability distribution:
\begin{align}
    p(\text{choose}~i) = p(x_k=i|x_{-k},c)
\end{align}
If satisfied, this process will converge to a stationary distribution matching $p(x|c)$.

We provided specific prompts that implement Gibbs Sampling with GPT-4 as follows: \textit{``You are a participant in a color judgment task. You will see an object and a color code in HSL format, however, one dimension of the given HSL color code is unknown. Your objective is to assign an apt integer to the unknown dimension to make the HSL color code match the given object as well as possible. Remember, it’s essential to complete the color, even if the generated color might seem unusual at times. Please limit your response to just the value you'd like to assign to the unknown dimension. Adjust the unknown dimension of HSL color to match the following object as well as possible: strawberry. Color: [270, 50, 'unknown']''}

\section{Recovering Color Representations from GPT-4}

To recover mental representations for a low-dimensional perceptual domain, color, we employed a variety of behavioral methods to engage with GPT-4. Most of these methods have been used to elicit human representations \parencite{sanborn_markov_2007, sanborn_uncovering_2010, harrison_gibbs_2020}. Hypothesizing that GPT-4 can mimic human decision-making processes, we substituted human participants with GPT-4, enabling us to harvest samples directly from the LLM's color representation.

\subsection{Stimuli}
Our experimental design mirrors the human study conducted by \textcite{harrison_gibbs_2020}, which used the HSL color values. Hue values range from 0 to 360, while both saturation and lightness extend from 0 to 100. We aimed to recover the representations of six specific objects within this color spectrum, enabling direct comparisons with corresponding human representations. These objects are `Chocolate', `Lemon', `Strawberry', `Grass', `Eggshell', and `Lavender', each offering a distinct color profile for analysis.

We adapted the human data from \textcite{harrison_gibbs_2020}. Their research demonstrated that an aggregated GSP method is particularly effective in eliciting color representations from human participants. To briefly describe the process, each aggregated GSP chain was randomly initialized with an HSL color. Participants manipulated one color dimension at a time using a slider. For each iteration, judgments from five participants were aggregated and their mean value was used as the seed for the next iteration. Participants were only allowed to participate in a given chain only once to ensure within-chain trial independence \parencite{harrison_gibbs_2020}.

In the experiment conducted by \textcite{harrison_gibbs_2020}, participants received the following instructions: ``In each trial of this study you will be presented with a word and a color and your task will be to modify that color using a slider such that it best matches the target word. No prior expertise is required to complete this task, just answer what you intuitively think is the right color.'' Participants then completed up to 20 trials, responding to the prompt: ``Adjust the slider to match the following word as well as possible: $\langle\text{word}\rangle$''.

\subsection{Procedure}
GPT-4 was assigned to tasks of recovering color representations for the six objects tested in \textcite{harrison_gibbs_2020} through Direct Prompting, Direct Sampling, MCMC, and Gibbs Sampling. Detailed descriptions of the implementation for each method, along with corresponding visual illustrations, are presented in Figure \ref{fig:methods}. For all six objects, we configured GPT-4's temperature at $1.0$. This setting makes the model's outputs based on the model's learned probabilities. To ensure robustness and reliability in our findings, we ran all four behavioral methods for a total of 500 iterations. For methods that are based on sampling algorithms (Direct Sampling, MCMC, and Gibbs Sampling), we reinitialized the methods four times. This results in a cumulative total of 2000 samples, which were generated across four distinct chains. Figure \ref{fig:percept} displays representative samples produced by each method, and Figure \ref{fig:sample} depicts the evolution of the mean color representation across successive iterations.

\section{Results}

\subsection{Convergence Diagnostic for Markov Chains}
\begin{figure}[t!]
    \centering
    \includegraphics[width=0.45\textwidth]{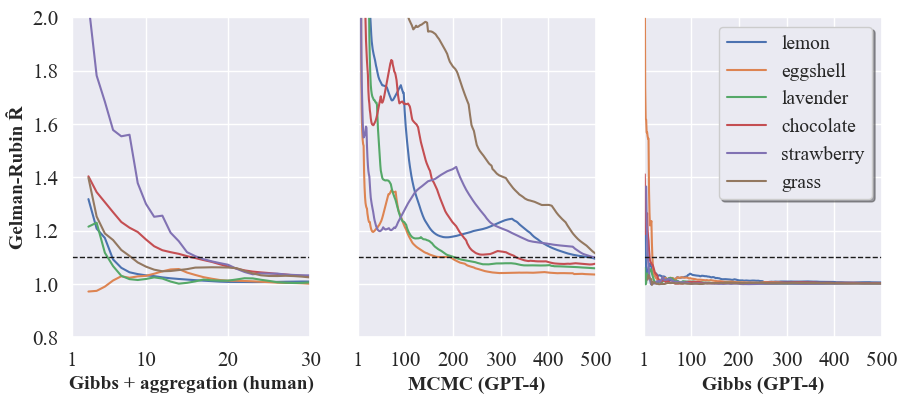}
    \caption{Cumulative $\hat{R}$ of Gibbs Sampling with People plus aggregation \textbf{(left)}, MCMC with GPT-4 \textbf{(middle)}, and Gibbs Sampling with GPT-4 \textbf{(right)}. Reaching the threshold of $1.1$ suggests convergence of the Markov chain.}
    \label{fig:rhat}
\end{figure}

The convergence of the Markov chains in MCMC and Gibbs sampling can be assessed using the Gelman-Rubin diagnostic \parencite{gelmanInferenceIterativeSimulation1992}. This diagnostic calculates the ratio of within-chain variance to between-chain variance, denoted as $\hat{R}$, serving as an indicator of the extent of convergence. A threshold of $\hat{R} \leq 1.1$ is commonly adopted as a criterion for satisfactory convergence in Markov chains. We present cumulative $\hat{R}$ values in Figure~\ref{fig:rhat}. In alignment with previous empirical studies involving human participants \parencite{harrison_gibbs_2020}, the MCMC with GPT-4 exhibited the slowest rate of convergence. In contrast, the Gibbs sampling method demonstrated significantly quicker convergence, typically reaching stability within 10 iterations.

\subsection{Representational Alignment of Humans and GPT-4}

\begin{table}[t!]
\begin{center} 
\caption{Distributional and mode distances (indicated in parentheses) between GPT-4 and human  representations. } 
\label{table:ce} 
\vskip 0.05in
\begin{tabular}[width=0.5\textwidth]{lcccc} 
\hline
    &  \makecell{Direct\\Prompting} & \makecell{Direct\\Sampling} & MCMC  &  \makecell{Gibbs\\Sampling} \\
\hline
Choc.  &  .99 (9.2) & .96 (4.6) & \textbf{.85} (\textbf{4.0}) & .95 (7.7) \\
Lemon  &  1.00 (13.5) & .99 (5.0) & \textbf{.95} (\textbf{3.2}) & 1.00 (9.3) \\
Strwb.  & \textbf{.80} (5.1) & .93 (5.5) & .93 (\textbf{4.7}) & .93 (9.1) \\
Grass  & 1.00 (6.3) & .99 (5.9) & \textbf{.98} (\textbf{5.5}) & .99 (6.5) \\
Eggsh.  &  .98 (3.7) & 1.00 (4.9) & .96 (5.7) & \textbf{.87} (\textbf{3.6}) \\
Lav.  &  1.00 (5.4) & .87 (3.8) & \textbf{.81} (5.8) & .97 (\textbf{3.6}) \\ 
\hline
\end{tabular} 
\end{center} 
\textit{Note.} Human representations for these objects were derived from data reported in \textcite{harrison_gibbs_2020}. Bold numbers represent the best correspondence with human among the four behavioral methods (smaller is better). 
From top to bottom, the tested objects are `Chocolate', `Lemon', `Strawberry', `Grass', `Eggshell', and `Lavender'. 
\end{table}

Upon verifying the convergence of the Markov chains, our analysis investigated the alignment of color representations between humans and GPT-4. First, the color space was discretized into a 18$\times$10$\times$10 grid across the H-S-L dimensions, adopting a broader bin width to minimize the impact of minor color variations. Next, we estimated the probability density within each defined bin. 

For the purpose of this comparison, we selected the human color representations derived from the GSP+aggregation condition reported in \textcite{harrison_gibbs_2020}, as this method has demonstrated best performance in recovering human mental representations.

We developed two metrics to evaluate the representational alignment between humans and GPT-4. The first metric aims to quantify the overall agreement between the two distributions, $\hat{p}_\text{human}(x|c)$ and $\hat{p}_\text{GPT-4}(x|c)$. For this purpose, we employed the Hellinger distance: 
\begin{align}
\hspace{-1mm}
    H^2(\hat{p}_\text{human}, \hat{p}_\text{GPT-4}) = \frac{1}{2}\sum_{dx\in\mathcal{X}} \Big(\sqrt{\hat{p}_\text{human}(dx)}- \sqrt{\hat{p}_\text{GPT-4}(dx)} \Big)^2 \nonumber
\end{align}
The Hellinger distance is symmetric and bounded between 0 and 1, where 0 indicates identical distributions and 1 indicates maximum dissimilarity. This bounded range can be more intuitive and easier to interpret than unbounded measures. It is more robust when dealing with distributions that have zero probabilities.

While assessing the overall distributional alignment is crucial, it is also important to examine the most probable or representative mental state (i.e., $\arg\max_x p(x|c)$). Accordingly, we measured the Euclidean distance between the modes of the mental representations as derived from GPT-4 and humans. This second metric allows for a focused comparison of the most probable representational in both representations.

\begin{figure}[t!]
    \centering    
    \includegraphics[width=0.5\textwidth]{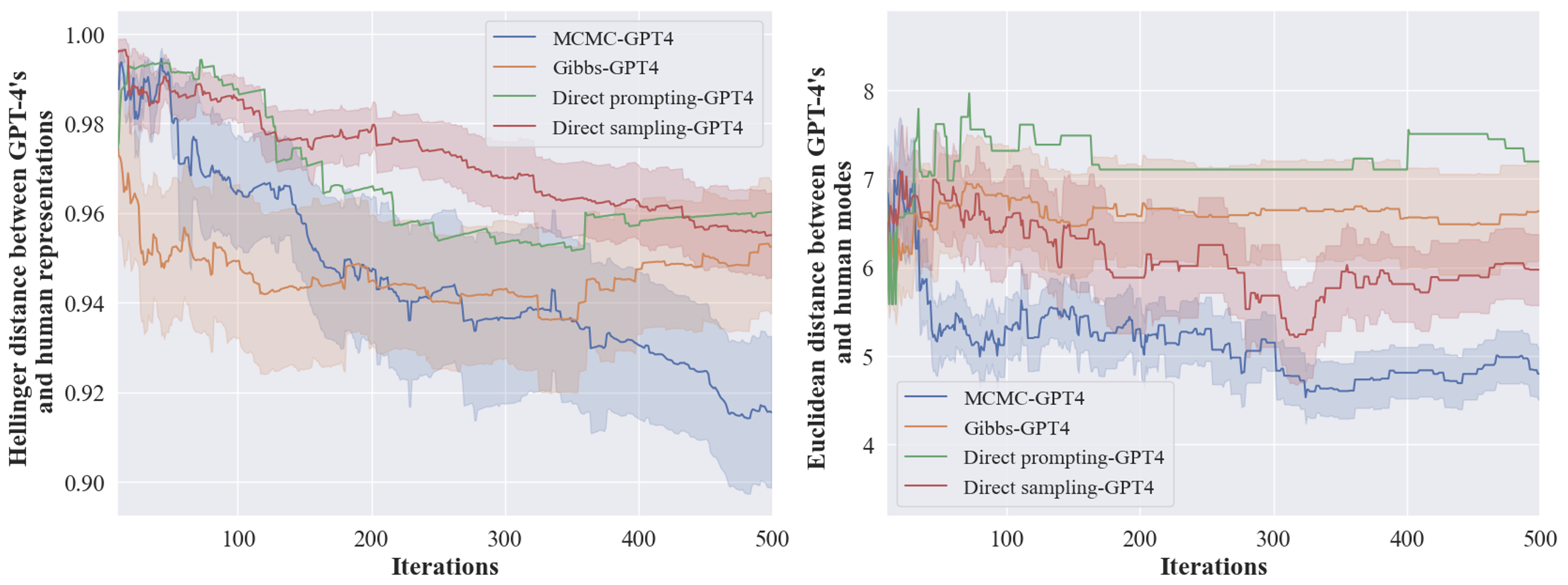}
    \caption{Comparing representations in humans and GPT-4. \textbf{(left)} Hellinger distance between the color representations derived from GPT-4 and those from humans. \textbf{(right)} Euclidean distances between the modes of representations from GPT-4 and humans. In both measures, lower numerical values are indicative of a stronger correspondence. Shaded areas indicate $\pm$SEM. }
    \label{fig:distance1}
\end{figure}

We calculated both metrics based on each of the 500-sample chains generated by Direct Sampling, MCMC, and Gibbs Sampling. Then these values were averaged over 4 repetitions of the sampling process. The resulting data are summarized in Table \ref{table:ce}. Moreover, the progression of representational alignment throughout these iterations is depicted in Figure \ref{fig:distance1}.

We found that among the methods employed, MCMC with GPT-4 exhibits a notably superior performance in closely approximating the overall distributions and the modes of most human color representations (see Table \ref{table:ce} and Figure \ref{fig:distance1}). Meanwhile, the other adaptive method, Gibbs Sampling with GPT-4, showed best performance in accurately representing eggshell and the mode of lavender. In contrast, both static methods, including Direct Prompting and Direct Sampling with GPT-4, significantly lag behind in performance. Overall, the integration of GPT-4 with adaptive methods was more efficient than the integration with static methods in replicating human color representations.

\section{Discussion}

We developed and evaluated a novel class of adaptive methods with LLMs. Our approach is grounded in two fundamental design principles: first, the incorporation of LLM outputs as integral components in sampling algorithms, and second, the dynamic modification of prompts based on previous responses from the LLMs. We tested integrating GPT-4 with various sampling algorithms, including Direct Sampling, MCMC, and Gibbs sampling. The objective was to recover human-like color representation. Our findings demonstrate that adaptive methods (MCMC and Gibbs sampling) significantly surpass the performance of static methods (Direct Prompting and Direct Sampling).

\subsection{Towards Doing Bayesian Inference with LLMs}

While we have focused on recovering conditional probabilities like $p(x|c)$ from GPT-4, the success of the methods we have presented here suggests that they could be adapted to sampling from other distributions. This capability is crucial in Bayesian inference, where many problems involve approximating the posterior probability of hypotheses $h$ given data $d$,  $p(h|d)$. This posterior probability is, in essence, a form of conditional probability distribution. Our methods could significantly broaden the scope for applying LLMs in Bayesian inference tasks. This can be achieved by constructing Markov chains with LLMs, which can be framed as simple as either choice-based or estimation-focused tasks.

The adaptive methods we employed are especially noteworthy. These methods dynamically alter prompts based on previous responses from LLMs, presenting a promising avenue for effectively conducting Bayesian inference. They not only simplify the task format for LLMs but also offer a more efficient means to navigate through the hypothesis space. While more advanced sampling algorithms such as Hamiltonian Monte Carlo \parencite{betancourt2017conceptual} and the No-U-Turn Sampler \parencite{hoffman2014no} could replace MCMC and Gibbs Sampling, the optimal choice of sampling algorithm should be determined by a combination of the target distribution's geometry and the response characteristics of the LLMs. This is because there are crucial assumptions that need to be satisfied (e.g., those outlined in Equations 2 and 3 for MCMC with GPT-4 and Gibbs sampling with GPT-4 respectively) to ensure that the sampling algorithms effectively converge to the correct target distribution.

It is important to note the distinction between our approach and other recent approaches for implementing Bayesian inference \textit{using} LLMs (e.g., \cite{wong2023word, zhang2023grounded}). For example, \textcite{wong2023word}'s proposal primarily leverages LLMs as translators, converting natural language inputs into probabilistic programming language statements. These statements are then subjected to Bayesian inference. This process essentially transforms LLMs into intermediaries, facilitating the translation from natural languages, which are inherently challenging for Bayesian inference, to symbolic representations that are more amenable to probabilistic programming languages, such as Church \parencite{goodman2012church}. In contrast, our proposal advocates for a more direct usage of LLMs in Bayesian inference, positioning them as the primary computational mechanism rather than mere translators. Our findings suggest that constructing a Markov chain \textit{with} LLMs for Bayesian inference might be more efficient compared to prompting LLMs directly.

\subsection{Limitations and Future Directions}

Our work highlights the potential of adaptive methods in revealing information about  LLMs. Nonetheless, it is imperative to conduct further research into how behavioral methods can be applied to LLM responses. The efficacy of adaptive methods is heavily contingent upon the congruence between our presupposed assumptions regarding the nature of LLM responses and the actual response patterns exhibited by these models. This alignment is critical for the successful implementation and optimization of adaptive methods.

Additionally, there are various hyperparameters within the sampling algorithms as well as the LLMs, such as the proposal distributions and the temperature, that offer opportunities for fine-tuning. Tailoring these parameters to specific domains could potentially enhance the performance of these algorithms and minimize the total number of token requests for LLMs. The adaptability and precision of such parameters are vital in harnessing the full potential of LLMs in specialized contexts. Our research paves the way for future explorations into optimizing these aspects to achieve greater efficiency and accuracy in applications of LLMs.

\printbibliography

@Article{sanborn_uncovering_2010,
    title = {Uncovering mental representations with {M}arkov chain {M}onte {C}arlo},
    volume = {60},
    pages = {63--106},
    number = {2},
    year = 2010, 
    journal = {Cognitive Psychology},
    author = {A. N. Sanborn and T. L. Griffiths and R. M. Shiffrin},
}

@Article{sanborn_markov_2007,
	title = {Markov Chain {M}onte {C}arlo with People},
	volume = {20},
	journal = {Advances in {N}eural {I}nformation {P}rocessing {S}ystems},
	author = {A. N. Sanborn and T. L. Griffiths},
	year = {2007},
}

@inproceedings{harrison_gibbs_2020,
	title = {Gibbs Sampling with People},
	volume = {33},
	pages = {10659--10671},
	booktitle = {Advances in {N}eural {I}nformation {P}rocessing {S}ystems},
	author = {Harrison, Peter and Marjieh, Raja and Adolfi, Federico and van Rijn, Pol and Anglada-Tort, Manuel and Tchernichovski, Ofer and Larrouy-Maestri, Pauline and Jacoby, Nori},
	year = 2020,
}

@article{gelmanInferenceIterativeSimulation1992,
  title = {Inference from {{Iterative Simulation Using Multiple Sequences}}},
  author = {Gelman, Andrew and Rubin, Donald B.},
  year = {1992},
  journal = {Statistical Science},
  volume = {7},
  number = {4},
  pages = {457--472},
  publisher = {{Institute of Mathematical Statistics}},
  issn = {0883-4237, 2168-8745},
  doi = {10.1214/ss/1177011136},
  urldate = {2024-01-14}
}

@article{barkerMonteCarloCalculations1965,
  title = {Monte {{Carlo Calculations}} of the {{Radial Distribution Functions}} for a {{Proton}}-{{Electron Plasma}}},
  shorttitle = {Monte {{Carlo Calculations}} of the {{Radial Distribution Functions}} for a {{Proton}}?},
  author = {Barker, A. A.},
  year = {1965},
  journal = {Australian Journal of Physics},
  volume = {18},
  number = {2},
  pages = {119--134},
  publisher = {{CSIRO PUBLISHING}},
  issn = {1446-5582},
  doi = {10.1071/ph650119}
}

@Book{torgerson1958theory,
  title={Theory and methods of scaling.},
  author={Torgerson, Warren S},
  year={1958},
  publisher={Wiley},
  address={New York}
}

@article{shepard1979additive,
  title={Additive clustering: Representation of similarities as combinations of discrete overlapping properties.},
  author={Shepard, Roger N and Arabie, Phipps},
  journal={Psychological Review},
  volume={86},
  number={2},
  pages={87--123},
  year={1979},
  publisher={American Psychological Association}
}

@article{rathje2023gpt,
  title={{GPT} is an effective tool for multilingual psychological text analysis},
  author={Rathje, Steve and Mirea, Dan-Mircea and Sucholutsky, Ilia and Marjieh, Raja and Robertson, Claire and Van Bavel, Jay J},
  year={2023},
  journal={PsyArXiv}
}

@article{geman1984stochastic,
  title={Stochastic relaxation, Gibbs distributions, and the Bayesian restoration of images},
  author={Geman, Stuart and Geman, Donald},
  journal={IEEE Transactions on pattern analysis and machine intelligence},
  number={6},
  pages={721--741},
  year={1984},
  publisher={IEEE}
}

@article{bubeck2023sparks,
  title={Sparks of artificial general intelligence: Early experiments with {GPT}-4},
  author={Bubeck, S{\'e}bastien and Chandrasekaran, Varun and Eldan, Ronen and Gehrke, Johannes and Horvitz, Eric and Kamar, Ece and Lee, Peter and Lee, Yin Tat and Li, Yuanzhi and Lundberg, Scott and others},
  journal={arXiv preprint arXiv:2303.12712},
  year={2023}
}

@article{martinTestingEfficiencyMarkov2012,
  title = {Testing the {{Efficiency}} of {{Markov Chain Monte Carlo With People Using Facial Affect Categories}}},
  author = {Martin, Jay B. and Griffiths, Thomas L. and Sanborn, Adam N.},
  year = {2012},
  journal = {Cognitive Science},
  volume = {36},
  number = {1},
  pages = {150--162}
}

@book{kay2009world,
  title={The world color survey},
  author={Kay, Paul and Berlin, Brent and Maffi, Luisa and Merrifield, William R and Cook, Richard},
  year={2009},
  publisher={CSLI Publications Stanford, CA}
}

@article{hoffman2014no,
  title={The {N}o-{U}-{T}urn sampler: {A}daptively setting path lengths in {H}amiltonian {M}onte {C}arlo.},
  author={Hoffman, Matthew D and Gelman, Andrew},
  journal={Journal Machine Learning Research},
  volume={15},
  number={1},
  pages={1593--1623},
  year={2014}
}

@article{betancourt2017conceptual,
  title={A conceptual introduction to {H}amiltonian {M}onte {C}arlo},
  author={Betancourt, Michael},
  journal={arXiv preprint arXiv:1701.02434},
  year={2017}
}

@article{wong2023word,
  title={From Word Models to World Models: Translating from Natural Language to the Probabilistic Language of Thought},
  author={Wong, Lionel and Grand, Gabriel and Lew, Alexander K and Goodman, Noah D and Mansinghka, Vikash K and Andreas, Jacob and Tenenbaum, Joshua B},
  journal={arXiv preprint arXiv:2306.12672},
  year={2023}
}

@article{goodman2012church,
  title={Church: a language for generative models},
  author={Goodman, Noah and Mansinghka, Vikash and Roy, Daniel M and Bonawitz, Keith and Tenenbaum, Joshua B},
  journal={arXiv preprint arXiv:1206.3255},
  year={2012}
}

@inproceedings{zhang2023grounded,
  title={Grounded physical language understanding with probabilistic programs and simulated worlds},
  author={Zhang, Cedegao E and Wong, Lionel and Grand, Gabriel and Tenenbaum, Joshua B},
  booktitle={Proceedings of the {A}nnual {C}onference of the {C}ognitive {S}cience {S}ociety},
  year={2023}
}

@article{lecun2015deep,
  title={Deep learning},
  author={LeCun, Yann and Bengio, Yoshua and Hinton, Geoffrey},
  journal={Nature},
  volume={521},
  number={7553},
  pages={436--444},
  year={2015},
  publisher={Nature Publishing Group UK London}
}

@article{vaswani2017attention,
  title={Attention is all you need},
  author={Vaswani, Ashish and Shazeer, Noam and Parmar, Niki and Uszkoreit, Jakob and Jones, Llion and Gomez, Aidan N and Kaiser, {\L}ukasz and Polosukhin, Illia},
  journal={Advances in {N}eural {I}nformation {P}rocessing {S}ystems},
  volume={30},
  year={2017}
}

@inproceedings{kornblith2019similarity,
  title={Similarity of neural network representations revisited},
  author={Kornblith, Simon and Norouzi, Mohammad and Lee, Honglak and Hinton, Geoffrey},
  booktitle={International {C}onference on {M}achine {L}earning},
  pages={3519--3529},
  year={2019},
  organization={PMLR}
}

@article{belinkov2022probing,
  title={Probing classifiers: Promises, shortcomings, and advances},
  author={Belinkov, Yonatan},
  journal={Computational Linguistics},
  volume={48},
  number={1},
  pages={207--219},
  year={2022},
  publisher={MIT Press One Broadway, 12th Floor, Cambridge, Massachusetts 02142, USA~…}
}

@article{zhang2023deep,
  title={Deep de Finetti: Recovering Topic Distributions from Large Language Models},
  author={Zhang, Liyi and McCoy, R Thomas and Sumers, Theodore R and Zhu, Jian-Qiao and Griffiths, Thomas L},
  journal={arXiv preprint arXiv:2312.14226},
  year={2023}
}

@article{alain2016understanding,
  title={Understanding intermediate layers using linear classifier probes},
  author={Alain, Guillaume and Bengio, Yoshua},
  journal={arXiv preprint arXiv:1610.01644},
  year={2016}
}

\end{document}